\documentclass[10pt,twocolumn,letterpaper]{article}

\usepackage{cvpr}
\usepackage{times}
\usepackage{epsfig}
\usepackage{graphicx}
\usepackage{amsmath}
\usepackage{amssymb}

\usepackage{exscale}
\usepackage[mathscr]{eucal}
\usepackage{bm}
\usepackage{eqlist} 
\usepackage{flushend}
\usepackage{lettrine}
\usepackage{algorithm}
\usepackage{algorithmic}
\usepackage{url}
\usepackage{multirow} 
\usepackage{booktabs} 
\usepackage{colortbl} 
\usepackage{lscape}
\usepackage{times} 
\usepackage{amsfonts}
\usepackage{psfrag}
\usepackage{latexsym}
\usepackage{eucal}
\usepackage[belowskip=0pt,aboveskip=5pt]{caption}

\usepackage{multirow} 
\usepackage{booktabs} 
\usepackage{colortbl} 
\usepackage{algorithmic}
\usepackage{algorithm}
\usepackage{color}
\usepackage{xfrac}

\newcommand{\+}[1]{\bm{#1}}  
\newcommand{\argmin}{\operatornamewithlimits{argmin}}

\newcommand{\Balpha}{\bm{\alpha}} 

\usepackage[pagebackref=true,breaklinks=true,letterpaper=true,colorlinks,bookmarks=false]{hyperref}

 \cvprfinalcopy 

\def\cvprPaperID{247} 

\ifcvprfinal\fi

\begin{document}

\title{Quality-based Multimodal Classification Using Tree-Structured Sparsity}

\author{Soheil Bahrampour\\
Pennsylvania State University\\
{\tt\small soheil@psu.edu}
\and
Asok Ray\\
Pennsylvania State University\\
{\tt\small axr2@psu.edu@psu.edu}
\and
Nasser M. Nasrabadi\\
Army Research Laboratory\\
{\tt\small nasser.m.nasrabadi.civ@mail.mil}
\and
Kenneth W.  Jenkins\\
Pennsylvania State University\\
{\tt\small jenkins@engr.psu.edu}
}
\maketitle

\begin{abstract}
Recent studies have demonstrated advantages of information fusion based on sparsity models for multimodal classification. Among several sparsity models, tree-structured sparsity provides a flexible framework for extraction of cross-correlated information from different sources and for enforcing group sparsity at multiple granularities. However, the existing algorithm only solves an approximated version of the cost functional and the resulting solution is not necessarily sparse at group levels. This paper reformulates the tree-structured sparse model for multimodal classification task. An accelerated proximal algorithm is proposed to solve the optimization problem, which is an efficient tool for feature-level fusion among either homogeneous or heterogeneous sources of information. In addition, a (fuzzy-set-theoretic) possibilistic scheme is proposed to weight the available modalities, based on their respective reliability, in a joint optimization problem for finding the sparsity codes. This approach provides a general framework for quality-based fusion that offers added robustness to several sparsity-based multimodal classification algorithms. To demonstrate their efficacy, the proposed methods are evaluated on three different applications -- multiview face recognition, multimodal face recognition, and target classification.
\end{abstract}

\section{Introduction}\label{sec:Introduction}

Information fusion using multiple sensors often results in better situation awareness and decision making~\cite{HL97}. While the information from a single sensor is generally localized and can be corrupted, sensor fusion provides a framework to obtain sufficiently local information from different perspectives, which is expected to be more tolerant to the errors of individual sources. Moreover, the cross-correlated information of (possibly heterogeneous) sources can be used for context learning and enhanced machine perception~\cite{WSSY02}.

Fusion algorithms are usually categorized into two levels: feature fusion~\cite{RG05} and classifier fusion~\cite{PBMS09,RG00}. Feature fusion methods combine various features extracted from different sources into a single feature set, which are then used for classification. On the other hand, classifier fusion aggregates decisions from several classifiers, where each classifier is built upon separate sources. While classifier fusion has been well studied, feature fusion is a relatively less-studied problem, mainly due to the incompatibility of feature sets~\cite{RKBT07}. A naive way of feature fusion is to concatenate features into a longer one~\cite{ZNZH11}, which may suffer from the curse of dimensionality. Moreover, the concatenated feature does not contain the cross-correlated information among the sources~\cite{RG05}. However, if above limitations are mitigated, feature fusion can potentially outperform the classifier fusion~\cite{KTR07}.

Recently, sparse representation has attracted the interest of many researchers, both for reconstructive and discriminative tasks~\cite{WYGSM09}. The underlying assumption is that if a dictionary is constructed with the training samples of all classes, only a few atoms of the dictionary, with the same label as the test data, should contribute to reconstruct the test sample. 
Feature level fusion using sparse representation has also been recently introduced and is often referred to as ``multi-task learning" in which the general goal is to represent samples jointly from several tasks/sources using different sparsity priors~\cite{SPNC13, SMJMHJ13, YZZW12}. In~\cite{NNT11}, a joint sparse model is introduced in which multiple observations from the same class are simultaneously represented by a few training samples. In other words, observations of a single scenario from different modalities would generate the same sparsity pattern of representing coefficients, which lies in a low-dimensional subspace. Similarly, modalities are fused with a joint sparsity model in~\cite{NNT11} and~\cite{SPNC13} for target classification and biometric recognition, respectively. Joint sparsity model relies on the fact that \textit{all} the different sources share the same sparsity patterns at atom level, which is not necessarily true and may limit its applicability. 

Another proposed solution is to group the relevant (correlated) tasks together and seek common sparsity within the group only~\cite{KGS11} or allowing small collaboration between different groups~\cite{KH12}. A more generalized approach is proposed in~\cite{KX09} for multi-task regression in which different tasks can be grouped in a tree-structured sparsity providing flexibility in fusion of different sources. Although the formulation of tree-structured sparsity proposed in~\cite{KX09} provides a framework to model different sparsity structures among multiple tasks, the proposed optimization algorithm only solves an approximation of the formulation and therefore cannot enforce the desired sparsity within different groups and sparsity can only be achieved within each task, separately. Moreover, in all the discussed multimodal fusion algorithms, including tree-structured sparsity, different modalities are assumed to have equal contributions for classification task. This can significantly limits the performance of fusion algorithms in dealing with occasional perturbation or malfunction of individual modalities. In~\cite{SPNC13}, a quality measure based on the joint sparse representation is introduced to quantify the quality of the data from different modalities. However, this index is measurable only after the sparse codes are obtained. Moreover, it measures the sparsity level of the representing coefficients which does not necessarily reflect the quality of individual modalities.

The major contributions of the paper are as follows:
($i$)~\emph{Reformulation and efficient optimization of the tree-structured sparsity for multimodal classification}: A finite number of separated problems are efficiently solved using the proximal algorithm~\cite{JMOB10} to provide an exact solution to the tree-structured sparse representation. The proposed learning facilitates feature level fusion of homogeneous/heterogeneous sources at multiple granularities.
($ii$)~\emph{Quality-based fusion}: A (fuzzy-set-theoretic) possibilistic approach~\cite{K86, KK93} is proposed to quantify the quality of different modalities in joint optimization with the reconstruction task. The proposed framework can be integrated with different sparsity priors (\eg joint sparsity or tree-structured sparsity) for quality-based fusion. The proposed method places larger weights on those modalities which have smaller reconstruction errors.
($iii$)~\emph{Improved performance for multimodal classification}: The improved performances and robustness of the proposed algorithms are illustrated on three applications -- multiview face recognition, multimodal face recognition, and target classification.

The rest of the paper is organized as follows. In Section~\ref{sec:SprseClass}, after briefly reviewing the joint sparsity model, multimodal tree-structured sparsity is reformulated and solved using the proximal algorithm. Section~\ref{sec:WeightedFusion} proposes the quality-based fusion which is followed by comparative studies in Section~\ref{sec:Results} and conclusions in Section~\ref{sec:Conclusions}.

\section{Multimodal sparse representation}\label{sec:SprseClass}
This section reviews the joint sparse representation classifiers and reformulates the tree-structured sparsity model~\cite{KX09} as a multimodal classifier. A proximal algorithm is then proposed to solve the associated optimization.

\subsection{Joint sparse representation classification}
Let $\mathcal{S} \triangleq \left\lbrace 1, \dots, S\right\rbrace$ be a finite set of available modalities used for multimodal classification and $C$ be the number of different classes in the dataset. Let the training data consist of $N = \sum_{c=1}^C N_c$ training samples from $S$ modalities, where $N_c$ is the number of training samples in the $c^{th}$ class. Let $n^s, s \in \mathcal{S},$ be the dimension of the feature vector for the $s^{th}$ modality and $\+x_{c,j}^s \in \mathbb{R}^{n^s}$ denote the $j^{th}$ sample of the $s^{th}$ modality belonging to the $c^{th}$ class, where $j\in \left\lbrace 1, \dots, N_c\right\rbrace$. In JSRC, $S$ dictionaries $\+X^s \triangleq [\+X_1^s \+X_2^s \dots \+X_C^s] \in \mathbb{R}^{n^s \times N}, s \in \mathcal{S},$ are constructed from the (normalized) training samples, where the class-wise sub-dictionary $\+X_c^s \triangleq \left[ \+x_{c,1}^s, \+x_{c,2}^s, \dots, \+x_{c,N_c}^s \right]  \in \mathbb{R}^{n^s \times N_c}$ consists of samples from the $c^{th}$ class and $s^{th}$ modality.

Given the test samples $\+y^s \in \mathbb{R}^{n^s}, s \in \mathcal{S},$ observed by $S$ different modalities from a single event, the goal is to classify the event. In the sparse representation classification, the key assumption is that a test sample $\+y^s$ from the $c^{th}$ class lies approximately within the subspace formed by the training samples of the $c^{th}$ class and can be approximated (or reconstructed) from \textit{a few} number of training samples in $\+X_c^s$~\cite{WYGSM09}. That is, if the test sample $\+y^s$ belongs to the $c^{th}$ class, it is represented as:
\begin{equation} \label{eq:TestRep}
\+y^s = \+{X^s\Balpha^s} + \+e,
\end{equation}
where $\Balpha^s \in \mathbb{R}^N$ is a coefficient vector whose entries are mostly $0$'s except for some of the entries associated with the $c^{th}$ class, i.e., $\Balpha^s = \left[ \+0^T, \dots, \+0^T, \Balpha_c^T, \+0^T, \dots, \+0^T \right]^T$, and $\+e$ is a small error term due to imperfectness of the samples.

In addition to the above assumption, JSRC enforces collaboration among different modalities to make a joint decision, where the coefficient vectors from different modalities are forced to have the same sparsity pattern. That is, the same training samples from different modalities are used to reconstruct the test data. The coefficient matrix $\+A = \left[\Balpha^1, \dots, \Balpha^S\right] \in \mathbb{R}^{N \times S}$, where $\Balpha^s$ is the sparse coefficient vector for reconstructing $\+y^s$, is recovered by solving the following $\ell_1/\ell_q$ joint optimization problem with $q \geq 1$:
\begin{equation} \label{eq:JSRC}
\argmin_{\+A= \left[\Balpha^1, \dots, \Balpha^S\right]} f(\+A) +\lambda\Vert \+A \Vert_{\ell_1/\ell_q}.
\end{equation}
In Eq.~(\ref{eq:JSRC}), $f (\+A) \triangleq \frac{1}{2}\sum_{s=1}^S\Vert \+y^s - \+{X^s\alpha^s}\Vert_{\ell_2}^2$ is the reconstruction error, $\ell_1/\ell_q$ norm is defined as $\Vert \+A \Vert_{\ell_1/\ell_q} = \sum_{j=1}^N \Vert \+a_j \Vert_{\ell_q}$ in which $\+a_j$'s are row vectors of $\+A$, and $\lambda>0$ is a regularization parameter. The number $q$ is usually set to $2$ and thus the second term in the cost function is refereed as $\ell_1/\ell_2$ penalty term. The above optimization problem encourages sharing of patterns across related observations so that the solution $\+A$ has a common support at the column level~\cite{NNT11}, which can be obtained by using different optimization algorithms (\eg alternating direction method of multipliers~\cite{YZ11}). The proximal algorithm is used in this paper~\cite{PB13}. 

Let $\delta_c(\alpha)\in \mathbb{R}^N$ be a vector indication function in which the rows corresponding to $c^{th}$ class are retained and the rest are set to zeros. The test data is classified using the class-specific reconstruction errors as:
\begin{equation} \label{eq:JSRCClassification}
c^* = \argmin_{c} \sum_{s=1}^S\Vert \+y^s - \+X^s\delta_c({\Balpha^s}^*)\Vert_{\ell_2}^2
\end{equation}
where ${\Balpha^s}^*$'s are optimal solutions of Eq.~(\ref{eq:JSRC}).

\subsection{Multimodal tree-structured sparse representation classification}\label{sec:MTSRC}
As discussed in Section~\ref{sec:Introduction}, although different sources are correlated, the joint sparsity assumption of JSRC may be too restrictive for some applications in which not all the different modalities are equally correlated and stronger correlations between some groups of the modalities may exist.

Tree-structured sparsity model provides a flexible framework to enforce prior knowledge in grouping different modalities by encoding them in a tree, where each leaf node represents an individual modality and each internal node represents a grouping of its child nodes. This arrangement allows modalities to be grouped at multiple granularity~\cite{KX09}. Adopting the definition in~\cite{JMOB10}, a tree-structured groups of modalities $\mathcal{G} \subseteq  \big(2^{\mathcal{S}} \ \setminus \ \emptyset \big)$ is defined as a collection of subsets of the set of modalities $\mathcal{S}$ such that $\bigcup_{g \in \mathcal{G}} g = \mathcal{S}$ and $\forall g, \tilde{g} \in \mathcal{G}, (g\cap \tilde{g}  \neq \emptyset ) \Rightarrow \left((g \subseteq \tilde{g}) \vee (\tilde{g} \subseteq g)\right)$. It is assumed here that $\mathcal{G}$ is ordered according to relation $\preccurlyeq$ which is defined as $(g \preccurlyeq \tilde{g}) \Rightarrow ((g \subseteq \tilde{g}) \vee (g \cap \tilde{g} =\emptyset))$. If the prior knowledge about grouping of modalities is not available, then hierarchical clustering algorithms could be used to find the tree structure~\cite{KX09}. 

Given a tree-structured collection $\mathcal{G}$ of groups, the proposed multimodal tree-structured sparse representation classification (MTSRC) is formulated as:
\begin{equation} \label{eq:MTSRC}
\argmin_{\+A= \left[\Balpha^1, \dots, \Balpha^S\right]} f(\+A) + \lambda \ \Omega \left(\+A\right)
\end{equation}
where $f(\+A)$ is defined the same as in Eq.~(\ref{eq:JSRC}), and the tree-structured sparse model is defined as:
\begin{equation} \label{eq:treeSparsity}
\Omega \left( \+A\right) \triangleq \sum_{j=1}^N\sum_{g \in \mathcal{G}} \omega_g \Vert \+a_{jg} \Vert_{\ell_2}.
\end{equation}
In Eq.~(\ref{eq:treeSparsity}), $\omega_g$ is a positive weight for group $g$ and $\+a_{jg}$ is a $(1\times S)$ row vector whose coordinates are equal to the $j^{th}$ row of $\+A$ for indices in the group $g$, and $0$ otherwise.

The above optimization problem allows sharing of patterns across related groups of modalities. Thus the optimal solution $\+A^*$ has a common support at the group level and the resulting sparsity is dependant on the relative weights $\omega_g$ of different groups~\cite{KX09}. 
Having obtained $\+A^*$, the test samples are classified using~(\ref{eq:JSRCClassification}). The tree-structured sparsity provides a flexible framework to enforce different sparsity priors. For example, if $\mathcal{G}$ consists of only one group, containing all modalities, then~(\ref{eq:MTSRC}) reduces to that of JSRC in~(\ref{eq:JSRC}). In another example where $\mathcal{G}$ consists of only singleton sets of individual modalities, no sparsity pattern is sought among different modalities and the optimization~(\ref{eq:MTSRC}) reduces to $S$ separate $\ell_1$ optimization problems.

\subsection{Optimization algorithm}\label{sec:Optimization}
As discussed in Section~\ref{sec:Introduction}, the optimization procedure proposed in~\cite{KX09} for tree-structured sparsity only solves an approximated version of the optimization problem~(\ref{eq:MTSRC}) and, therefore, does not necessarily results in a solution with desired group sparsity. In this section, an accelerated proximal gradient method~\cite{PB13} is used to solve~(\ref{eq:MTSRC}) in which the optimal solution is obtained by solving a finite number of tractable optimization problems without approximating the cost function. Let the initial value of $\+A$, which can be chosen as an arbitrary sparse vector, be zero. Then, at $k^{th}$ iteration, the proposed accelerated proximal gradient is as follows~\cite{PB13}:
\begin{equation}\label{eq:AccPro}
\begin{aligned}
\+B^{k+1} &= \+{\hat{A}}^{k} + \rho^k \left( \+{\hat{A}}^{k}- \+{\hat{A}}^{k-1}\right)\\
\+{\hat{A}}^{k+1} &= \operatorname{prox}_{\lambda t^k \Omega}\left( \+B^{k+1}-t^k \nabla f\left( \+{\+B^{k+1}}\right) \right) 
\end{aligned}
\end{equation}
where $t^k$ is the step size at time step $k$ which can be set as a constant or be updated using a line search algorithm~\cite{PB13}; $\+{\hat{A}}^{k}$ is the estimation of the optimal solution $\+A$ at time step $k$; the extrapolating parameter $\rho^k$ could be chosen as $\frac{k}{k+3}$; and the associated proximal optimization problem is defined as:
\begin{equation}\label{eq:proximal}
\operatorname{prox}_{\beta \Omega } \left(\+V\right)= \argmin_{\+U \in \mathbb{R}^{N \times S}} \Omega \left(\+U\right) + \frac{1}{2\beta}\Vert \+U - \+V \Vert_{F}^2,
\end{equation}
where $\Vert . \Vert_{F}$ is the Frobenius norm . Using Eq.~(\ref{eq:treeSparsity}), the proximal optimization problem is reformulated as:
\begin{align}\label{eq:separateProximal}
&\operatorname{prox}_{\beta \Omega } \left(\+V\right) = \notag\\ &\argmin_{\+U \in \mathbb{R}^{N \times S}} \sum_{j=1}^N\left(\sum_{g \in \mathcal{G}} \omega_g \Vert \+u_{jg} \Vert_{\ell_2} + \frac{1}{2\beta} \Vert \+u_j - \+v_j\Vert_{\ell_2}^2\right)
\end{align}
where $\+u_{jg}$ is defined similar to $\+a_{jg}$ in Eq.~(\ref{eq:treeSparsity}); and $\+u_j$ and $\+v_j$ are the $j^{th}$ rows of $\+U$ and $\+V$, respectively. Consequently, the solution of (\ref{eq:separateProximal}) is obtained by $N$ separate optimizations on $S$-dimensional vectors. Since the groups are defined to be ordered, each of the optimization problems can be solved in a single iteration using the dual form~\cite{JMOB10}. Therefore, the proximal step of the tree-structured sparsity can be solved with the same computational cost as that of joint sparsity. Algorithm~\ref{alg:Proximal}, which is a direct extension of the optimization algorithm in~\cite{JMOB10}, solves the proximal step. It should be noted that the computational complexity of the optimization algorithm grows linearly as the number of training samples increases. One can potentially learn the dictionaries with (significantly) fewer number of atoms using dictionary learning algorithms~\cite{KW12}. This paper uses the Sparse Modeling Software~\cite{JMOB10} to solve the proximal step.

\begin{algorithm}[!t]
\footnotesize 
\caption{\small Algorithm to solve the proximal optimization step (Eq.~(\ref{eq:separateProximal})) of the accelerated proximal gradient method (Eq.~(\ref{eq:AccPro})) corresponding to the MTSRC optimization problem (Eq.~(\ref{eq:MTSRC}))}\label{alg:Proximal}
\begin{algorithmic}[1]
	\REQUIRE $V \in \mathbb{R}^{N \times S}$, ordered set of groups $\mathcal{G}$, weights $\omega_g$ for each group $g \in \mathcal{G}$, and scaler $\beta$.
	\ENSURE $U \in \mathbb{R}^{N \times S}$ 
        \FOR{$j= 1, \dots, N$}
        	\STATE Let $\+{\eta}^1= \dots= \+{\eta}^{|\mathcal{G}|}=\+0$ and $\+u_j = \+v_j$.
        	\FOR{$g = g_1, g_2, \dots \in \mathcal{G}$}
        		\STATE $\+u_j = \+v_j - \sum_{h \neq g} \+{\eta}^h.$
        		\STATE $\+{\eta}^g = \left\{\begin{array}{ll} \frac{\+u_{jg}}{\Vert \+u_{jg} \Vert_{\ell_2}}, & \textrm{ if } \Vert \+u_{jg} \Vert_{\ell_2} > \beta \omega_g \\
        		\+u_{jg}, & \textrm{ if } \Vert \+u_{jg} \Vert_{\ell_2} \leq \beta \omega_g
        		\end{array}\right. .$
        	\ENDFOR	
        \STATE $\+u_j = \+v_j - \sum_{g \in \mathcal{G}}\+{\eta}^h.$
        \ENDFOR
\end{algorithmic}
\end{algorithm}

\section{Weighted scheme for quality-based fusion}\label{sec:WeightedFusion}
In most of the sparsity-based multimodal classification algorithms, including JSRC and MTSRC, it is inherently assumed that available modalities contribute equally. This may significantly limit the performance in dealing with occasional perturbation or malfunction of individual sources. Ideally, a fusion scheme should \textit{adaptively} weight the modalities based on their reliabilities. In~\cite{SPNC13}, a quality measure is introduced for JSRC, where a sparsity concentration index is calculated to quantify the quality of modalities. The main limitation of this approach, however, is that the index is obtained only after the sparse codes are calculated and a weak modality may hurt the performances of other modalities due to the enforced sparsity priors. Moreover, the index is defined based on the sparsity levels and does not necessarily reflect the quality of each modalities. This paper proposes to find the adaptive quality of each modality and sparse codes jointly in a single optimization problem.

Let $\mu^s$ be the quality weight for the $s^{th}$ modality. A weighted scheme for multimodal reconstruction, with similar structure to Eq.~(\ref{eq:MTSRC}), is proposed as follows:
\begin{equation}\label{eq:weightedScheme}
\argmin_{\+A= \left[\Balpha^1, \dots, \Balpha^S\right], \mu^s} \sum_{s=1}^S \frac{{\left( \mu^s\right)}^m}{2}\Vert \+y^s - \+{X^s\alpha^s}\Vert_{\ell_2}^2 + \Psi\left(\+A\right),
\end{equation}
with the constraint $\mu^s \geq 0, \ \forall s\in\mathcal{S}$, where the exponent $m \in (1, \infty)$ is a fuzzifier parameter, similar to formulation of fuzzy c-means clustering~\cite{BEF84}; and  $\Psi\left(\+A\right)$ enforces desired sparsity priors within the modalities (\eg $\ell_1/\ell_2$ constraint in JSRC or tree-structured sparsity prior of MTSRC). 

Another constraint on $\mu^s$ is necessary to avoid a degenerate solution of Eq.~(\ref{eq:weightedScheme}). A constraint such as $\sum_{s=1}^S \mu^s = 1$ is apparently feasible; however, since $m>1$ in Eq,~(\ref{eq:weightedScheme}), the larger weight of a modality compared to those of other modalities may effectively increase this weight close to $1$ while forcing the rest of the weights toward $0$.  To alleviate this problem, a ``possibility"-like constraint, similar to the possibilistic fuzzy $c$-means clustering~\cite{BMS11, KK93}, is proposed to allow the weights of different modalities to be specified independently. The proposed composite optimization problem to achieve quality-based multimodal fusion is posed as:
\begin{align} \label{eq:WeightedScheme}
&\argmin_{\+A=  \left[\Balpha^1, \dots, \Balpha^S\right], \mu^s} \left(\sum_{s=1}^S \frac{{\left( \mu^s\right)}^m}{2}\Vert \+y^s - \+{X^s\alpha^s}\Vert_{\ell_2}^2 + \Psi\left(\+A\right) + \right. \notag\\ & \left. \sum_{s=1}^S\lambda_{\mu^s}\left( 1- \mu^s \right)^m\right)  , \mu^s \geq 0, \forall s \in \mathcal{S}
\end{align}
where $\lambda_{\mu^s}$ are the regularization parameters for $s \in \mathcal{S}$.
After finding optimal $(\mu^s)^*$ and $\+A^*$, the test samples are classified using the weighted reconstruction errors, i.e.,
\begin{equation} \label{eq:WeightedClassification}
c^* = \argmin_{c} \sum_{s=1}^S\left( \mu^s\right)^m\Vert \+y^s - \+X^s\delta_c({\Balpha^s}^*)\Vert_{\ell_2}^2.
\end{equation}

The optimization problem in~(\ref{eq:WeightedScheme}) is not jointly convex in $\+A$ and $\mu^s$. A sub-optimal solution can be obtained by alternating between the updates of $\+A$ and $\left\lbrace \mu^s\right\rbrace$, minimizing over one variable while keeping the other variable fixed. The accelerated proximal gradient algorithm in~(\ref{eq:AccPro}) with $f (\+A) = \frac{1}{2}\sum_{s=1}^S {\left( \mu^s\right)}^m\Vert \+y^s - \+{X^s\alpha^s}\Vert_{\ell_2}^2$ is used to optimize over $\+A$ and optimal $\left\lbrace \mu^s\right\rbrace$ at each iteration of the alternative optimization  is found in a closed form~\cite{KK93} as:
\begin{equation} \label{eq:UpdateWeights}
\mu^s = \frac{1}{1+ {\left(\frac{\Vert \+y^s - \+X^s\Balpha^s\Vert_{\ell_2}^2}{\lambda_{\mu^s}} \right)}^{\frac{1}{m-1}} }, s \in \mathcal{S}.
\end{equation}
The regularization parameters $\lambda_{\mu^s}$ need to be chosen based on the desired bandwidth of the possibility distribution for each modality. In this paper, optimization over $\+A$ is first performed without weighting scheme to find an initial estimate of $\+A$. Also the following definition, similar to possibilistic clustering~\cite{BMS11}, is used to determine and fix $\lambda_{\mu^s}$:
\begin{equation} \label{eq:WeightRegul}
\lambda_{\mu^s} = \Vert \+y^s - \+X^s\Balpha^s\Vert_{\ell_2}^2,
\end{equation}
resulting all $\mu^s$ to be 0.5 initially. It is observed that only a few iterations is required for the algorithm to converge. In this paper, the number of alternations and the fuzzifier $m$ are set to be 10 and 2, respectively. Algorithm~\ref{alg:WeightedFusion} summarizes the proposed quality-based multimodal fusion method. As discussed, this scheme can be used with different sparsity priors as long as optimal $\+A$ can be found efficiently~\cite{PB13}.

\begin{algorithm}[!h]
\footnotesize 
\caption{\small Quality-based multimodal fusion.}\label{alg:WeightedFusion}
\begin{algorithmic}[1]
	\REQUIRE Initial coefficient matrix $\+A$, dictionary $\+X^s$ and test sample $\+y^s$ of the $s^{th}$ modality, $ s\in \left\lbrace 1, \cdots, S \right\rbrace$, and number of iterations $T$.
	\ENSURE Coefficient matrix $\+A$ and weights $\mu^s$ as a solution to Eq.(\ref{eq:weightedScheme}).
	    \STATE Set $\lambda_{\mu^s}$ using Eq.~(\ref{eq:WeightRegul})
        \FOR{$k= 1, \dots, T$}
        	\STATE Update weights $\mu^s$ using Eq.~(\ref{eq:UpdateWeights})
        	\STATE Update $\+A$ by solving Eq.~(\ref{eq:weightedScheme}) with fixed $\mu^s$.
        \ENDFOR
\end{algorithmic}
\end{algorithm}

\section{Results and discussion}\label{sec:Results}
In this section we present the results for the proposed MTSRC and weighted scheme in three different applications:  multiview face recognition, multimodal face recognition, and multimodal target classification. For MTSRC, the group weights $\omega_g$ are selected using a combination of aprior information/assumption and cross validation. We assumed equal weights for the same sized groups which reduces the number of tuning parameters. The relative weights between the groups with different sizes, however, are not fixed and are selected using cross-validation in a finite set $\lbrace 10^{-5},10^{-4},...,10^{5}\rbrace$. Hierarchical clustering can also be used to tune the weights~\cite{KX09}. It is observed that bigger groups are usually assigned with bigger weights in the studied applications. Thus MTSRC intuitively enforces collaboration among all the modalities and yet provides flexibility (compared to JSRC) by allowing collaborations to be formed at lower granularities as well. It is observed in all the studied applications that MTSRC performs similarly when the weights are varied without being reordered. 

For the weighted scheme, JSRC and MTSRC are equipped with the modality weights and the resulting algorithms are denoted as JSRC-W and MTSRC-W, respectively. The performance of the proposed feature-level fusion algorithms is compared with that of several state-of-the-art decision-level and feature-level fusion algorithms. For decision-level fusion, outputs of the independent classifiers, each trained on separate modality, are aggregated by adding the corresponding probability outputs of each modality, which is denoted as \textit{Sum}. For this purpose, SVM~\cite{B06}, sparse representation classifier (SRC)~\cite{WYGSM09}, and sparse logistic regression(SLR)~\cite{KCFH05} are used. The proposed methods are also evaluated using existing feature-level fusion methods that include holistic sparse representation classifier (HSRC)~\cite{ZNZH11}, JSRC~\cite{NNT11}, joint dynamic sparse representation classifier (JDSRC)~\cite{ZNZH11} and relaxed collaborative representation (RCR)~\cite{YZZW12}. 

\begin{figure}[!htbp]
   \centering
         \includegraphics[scale=0.3]{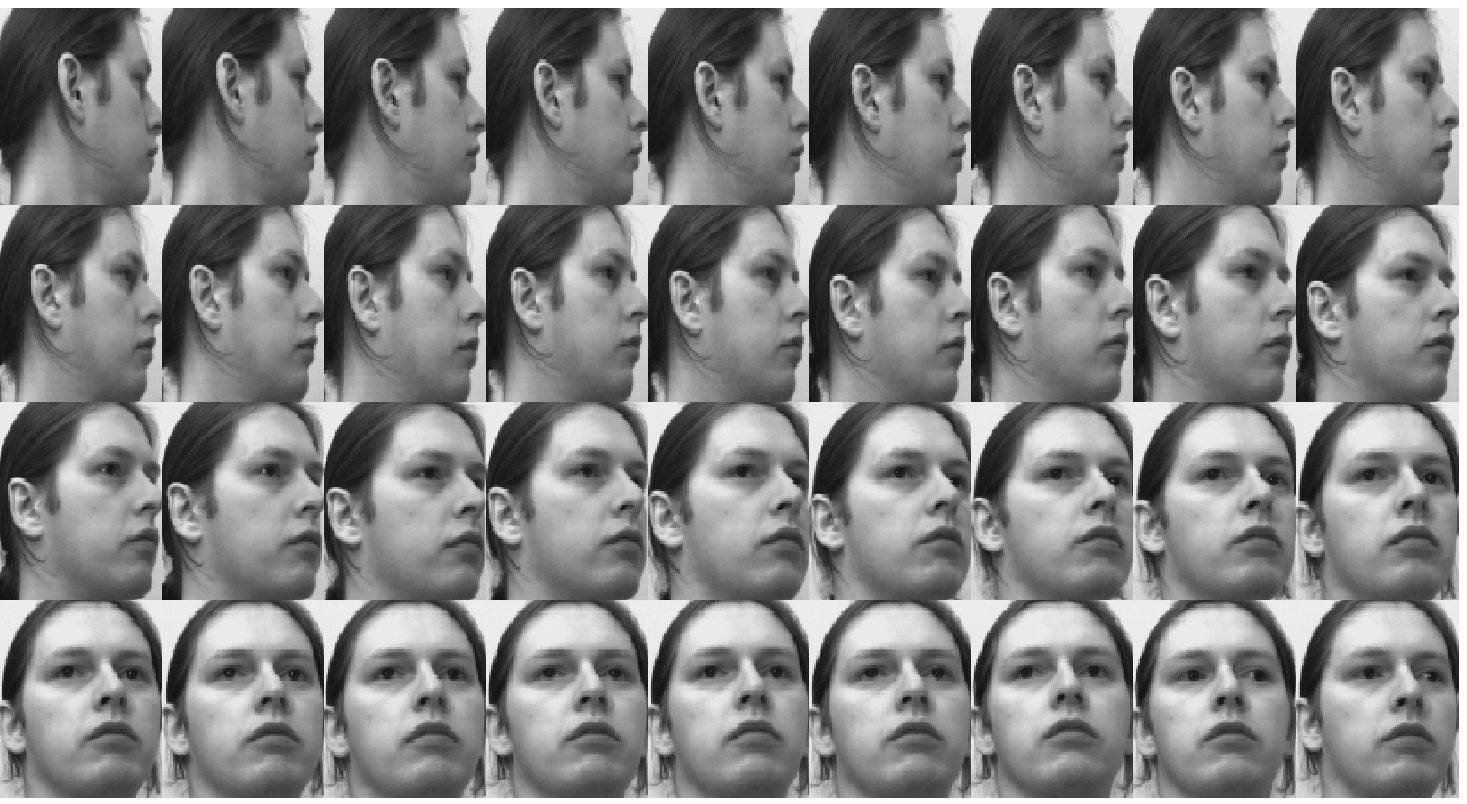}
         \caption{Different poses for one individual in the UMIST database that is divided into four different view-ranges shown by four rows.}
      \label{fig:umist}
\end{figure}

\begin{figure}[!htbp]
   \centering
         \includegraphics[scale=0.35]{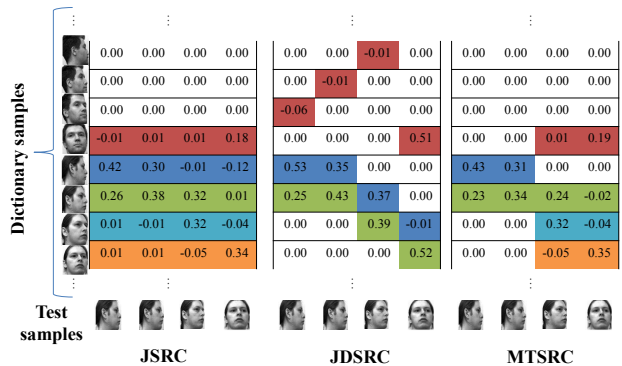}
         \caption{Representation coefficients generated by JSRC, JDSRC, and MTSRC on test observations that correspond to $4$ different views on the UMIST database. JSRC enforces joint sparsity among all views and requires the same sparsity pattern at atom level. JDSRC allows contributions from different training samples to approximate a set of test observations and requires the same sparsity pattern at class level. MTSRC enforces joint sparsity only when relevant.}
      \label{fig:SparseCofUmist}
\end{figure}

\subsection{Multiview face recognition}\label{sec:UMIST}
In this experiment, we evaluate the performance of the proposed MTSRC for multiview face recognition using UMIST face database which consists of 564 cropped images of 20 individuals with mixed race and gender~\cite{GA98}. Poses of each individual are sorted from profile to frontal views and then divided into $S$ view-ranges with equal number of images in each view-range. This facilitates multiview face recognition using UMIST database. The performance of the algorithms are tested using different values of view-ranges $S \in \lbrace2,3,4\rbrace$. It should be noted that the environment is unconstrained and captured faces may have pose variations within each view-range. Different poses for one of the individual in the UMIST database is shown in Fig.~\ref{fig:umist} where the images are divided into four view-ranges, shown in four rows. Due to limited number of training samples, a single dictionary is constructed by randomly selecting one (normalized) image per view for all the individuals in the database which is shared among different view-ranges. The rest of the images are used as the test data.

As observations from closeby views are more likely to share similar poses and be correlated, the tree structured sparsity of MTSRC is chosen to enforce group sparsity within related views. For this purpose, the tree-structured sets of groups using 2, 3, or 4 view-ranges are selected to be $\lbrace \lbrace 1 \rbrace, \lbrace 2 \rbrace, \lbrace 1,2 \rbrace \rbrace,$ $\lbrace \lbrace 1 \rbrace, \lbrace 2 \rbrace, \lbrace 3 \rbrace,\lbrace 1,2,3 \rbrace \rbrace,$ $\lbrace \lbrace 1 \rbrace, \lbrace 2 \rbrace, \lbrace 1,2 \rbrace, \lbrace 3 \rbrace, \lbrace 4 \rbrace, \lbrace 3,4 \rbrace, \lbrace 1,2,3,4 \rbrace \rbrace$, respectively. Fig.~\ref{fig:SparseCofUmist} shows the representation coefficients generated by JSRC, JDSRC, and MTSRC on a test scenario where four different view-ranges are used. As expected, JSRC enforces joint sparsity among all different views at atom level, which may be too restrictive due to relatively less correlation between frontal and profile views. JDSRC relaxes joint sparsity constraint at atom level and allows contributions from different training samples to approximate one set of the test observations but still requires joint sparsity pattern at class level. As shown, proposed MTSRC enforces joint sparsity within relevant views and also among all modalities and has the most flexible structure for multimodal classification. The results of 10-fold cross validation using different sparsity priors are summarized in Table~\ref{tab:UMISTResults}. As seen, the MTSRC method achieves the best performance. Since the quality of different view-ranges are similar, JSRC-W and MTSRC-W result in similar performances as those of JSRC and MTSRC, respectively, and therefore are omitted here.

\begin{table}[!t] 
\caption{Multiview classification results obtained on the UMIST database.}
\label{tab:UMISTResults}
\centering
\small
\begin{tabular}{lcccc}
 & 2 Views & 3 Views & 4 Views & Avg.\\
\toprule
HSRC~\cite{ZNZH11}  & 84.37 & 97.80 & 99.91 & 94.03 \\
JSRC~\cite{NNT11}  & 87.77 & 99.51 & 99.91 & 95.73\\
JDSRC~\cite{ZNZH11} & 86.52 & 98.96 & 99.91 & 95.13\\
MTSRC & \textbf{ 88.42} & \textbf{99.63} & \textbf{100.00} & \textbf{96.02} \\
\bottomrule
\end{tabular}
\end{table} 

\begin{figure}[!t]
   \centering
         \includegraphics[scale=0.18]{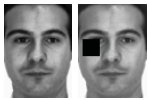}
         \caption{A test image and its occlusion in the AR dataset.}
      \label{fig:AR_samples}
\end{figure}

\begin{table}[!t] 
\caption{Correct classification rates obtained using individual modalities in AR database. Modalities include left and right periocular, nose, mouth, and face.}
\label{tab:ARIndResults}
\centering
\small
\setlength{\tabcolsep}{2mm}
\begin{tabular}{cccccc}
 & L periocular & R periocular & Nose & Mouth & Face \\
\toprule
SVM & 71.00 & 74.00 & 44.00 & 44.29 & 86.86 \\
SRC & 79.29 & 78.29 &  63.43 & 64.14 & 93.71\\
\bottomrule
\end{tabular}
\end{table} 

\begin{table}[!t] 
\caption{Multimodal classification results obtained on the AR database.}
\label{tab:ARFusResults}
\centering
\small
\begin{tabular}{lclc}
Method & CCR & Method & CCR \\
\toprule
SVM-Sum~\cite{SPNC13} & 92.57 & SLR-Sum~\cite{SPNC13} & 77.9 \\
JDSRC~\cite{ZNZH11} & 93.14 & RCR~\cite{YZZW12} & 94.00 \\
JSRC~\cite{NNT11} & 95.57 & JSRC-W & 96.43 \\
MTSRC & 97.14 & MTSRC-W & 97.14 \\
\bottomrule
\end{tabular}
\end{table} 

\begin{figure}[!t] %
   \centering
    \includegraphics[scale=0.16]{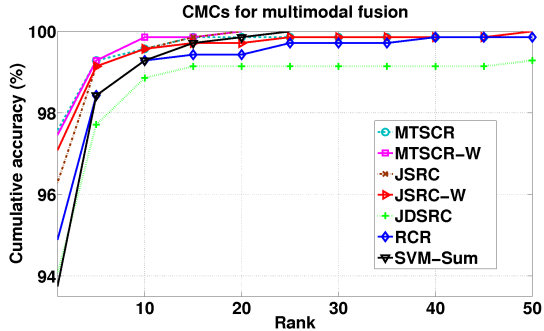}
    \caption{CMCs obtained using multimodal classification algorithms on AR database with random occlusion.}
      \label{fig:CMC_AR}
\end{figure}

\subsection{Multimodal face recognition: AR database}\label{sec:AR}

In this set of experiments, the performance of the proposed algorithms are evaluated on the AR database (Figure~\ref{fig:AR_samples}) which consists of faces under different poses, illumination and expression conditions, captured in two sessions~\cite{MB98}. A set of 100 users are used, each consisting of seven images from the first session as the training samples and seven images from the second session as test test samples. A small randomly selected portion of training samples, 30 out of 700, is used as validation set for optimizing the design parameters of classifiers. Fusion is taken on five modalities, similar to setup in~\cite{SPNC13}, including left and right periocular, nose, and mouth regions, as well as the whole face. After resizing the images, intensity values are used as features for all modalities. Results of using individual modalities for classification using SVM and SRC algorithms are shown in Table~\ref{tab:ARIndResults}. As expected, the whole face is the strongest modality in the sense of solving the identification task followed by left and right eyes. For MTSRC, the groups are chosen to be $\mathcal{G} = \lbrace g_1, g_2, g_3, g_4, g_5, g_6, g_7\rbrace= $ $\lbrace \lbrace 1 \rbrace, \lbrace 2 \rbrace, \lbrace 1,2 \rbrace, \lbrace 3 \rbrace, \lbrace 4 \rbrace, \lbrace 5 \rbrace, \lbrace 1,2,3,4,5 \rbrace \rbrace$, where $1$ and $2$ represents left and right periocular and $3,4,5$ represents other modalities. Weights are selected using a similar approach discussed in Section~\ref{sec:Border} to encourage group sparsity between all modalities and also joint representation for left and right periocular in lower granularity. The performances of the proposed algorithms are compared with several competitive methods as shown in Table~\ref{tab:ARFusResults}. Fig.~\ref{fig:CMC_AR} shows the corresponding cumulative matched score curves (CMC). CMC is a performance measure, similar to ROC, which is originally proposed for biometric recognition systems~\cite{BCPRS05}. As shown, the tree-structured sparsity based algorithms achieve the best performances with correct classification rate (CCR) of $97.14\%$. 

\begin{figure}[!t] %
   \centering
    \includegraphics[scale=0.16]{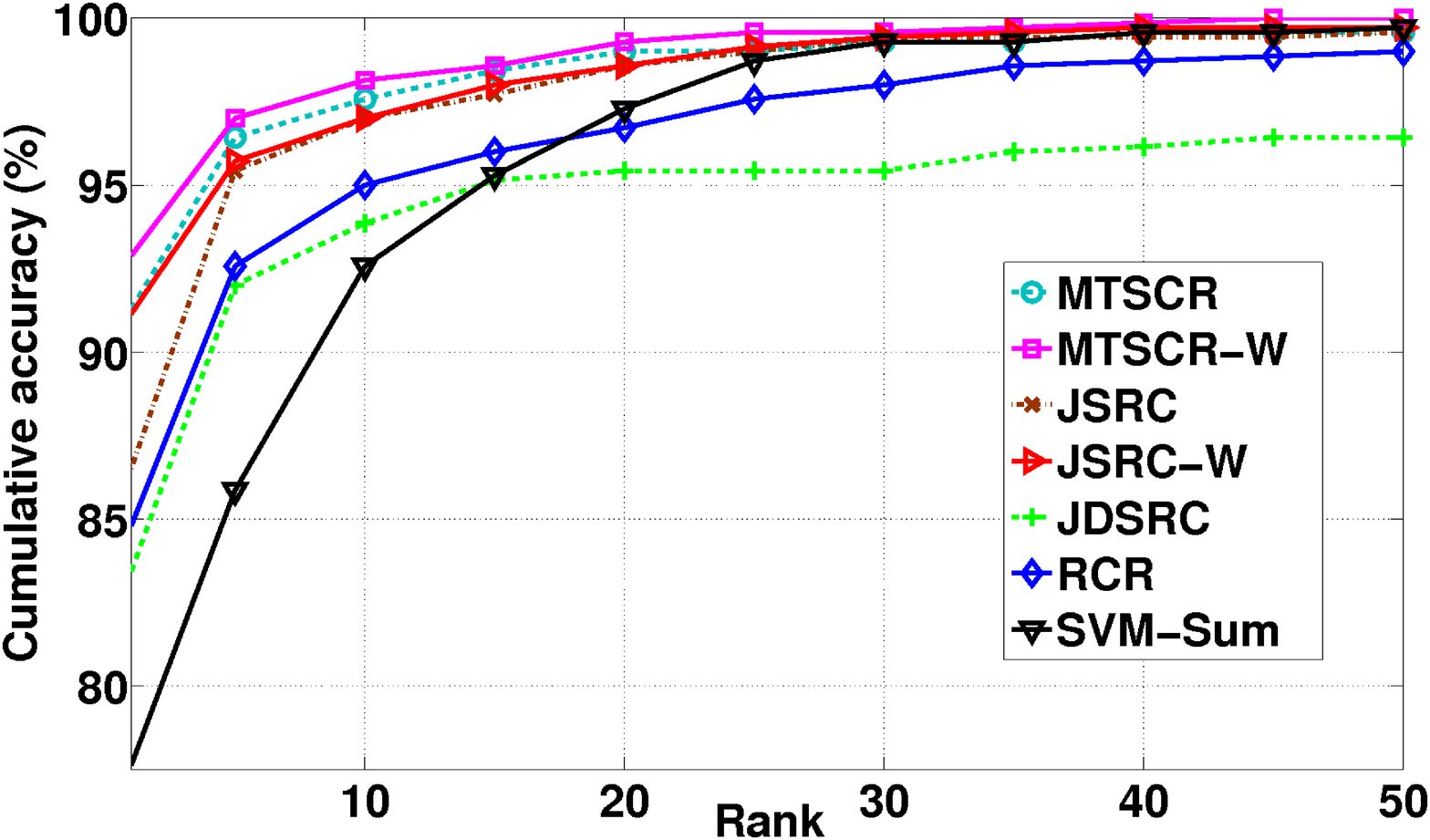}
    \caption{CMCs obtained using different multimodal classification algorithm on AR database with random occlusion.}
      \label{fig:CMC_AR_Block15}
\end{figure}

\begin{figure}[!t]
   \centering
         \includegraphics[scale=0.16]{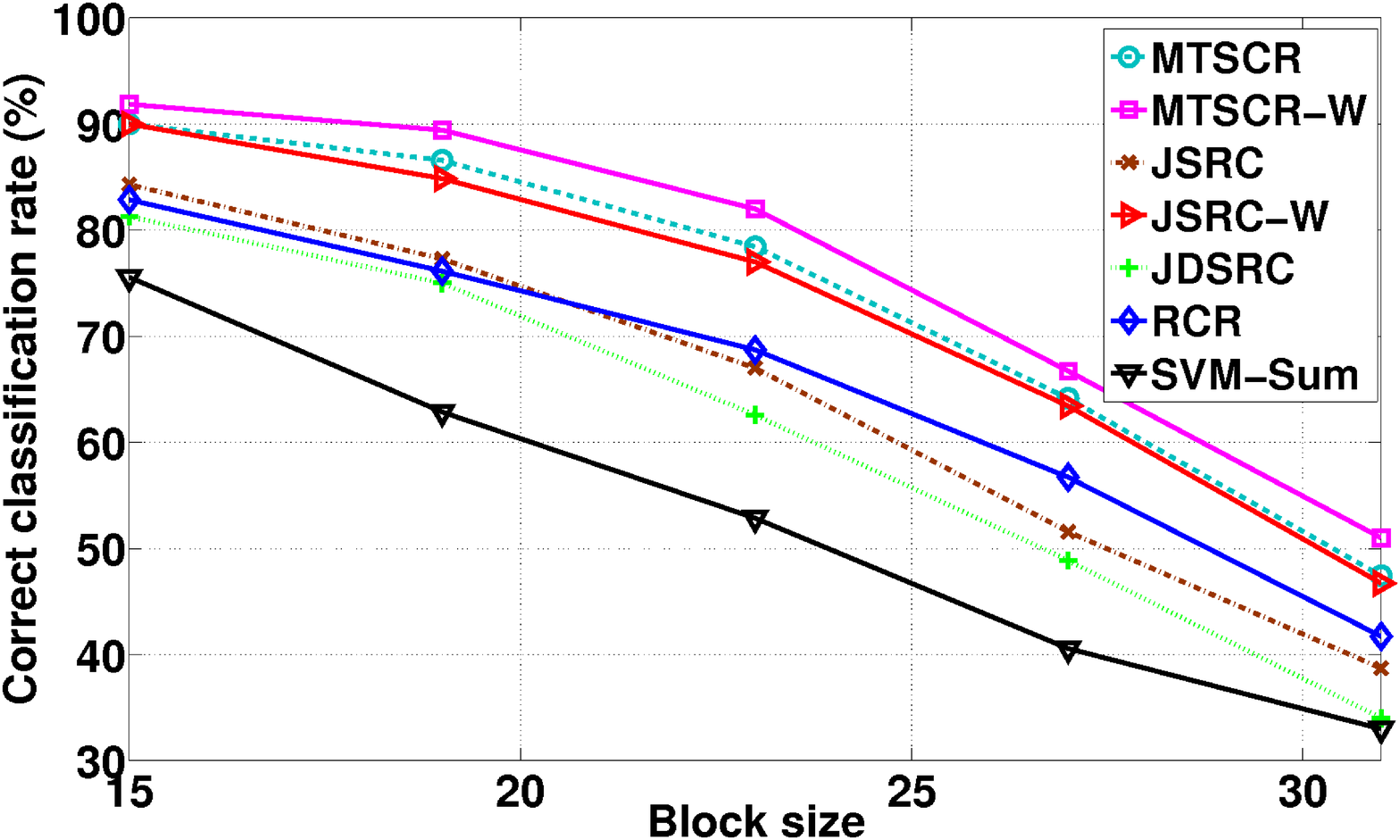}
         \caption{Correct classification rates of multimodal classification algorithms with block occlusion of different sizes.}
      \label{fig:AR_BlockSizes}
\end{figure}

To compare the robustness of different algorithms, each test images is occluded by randomly chosen block (See Fig.~\ref{fig:AR_samples}). Fig.~\ref{fig:CMC_AR_Block15} shows the CMC's generated when the size of occluding blocks are 15. As seen, the proposed tree-structured algorithms are the top performing algorithms. 
Fig.~\ref{fig:AR_BlockSizes} compares CCR of different algorithms with block occlusion of increasing sizes. MTSRC-W has the most robust performance. Also it is observed that the weighted scheme significantly improves the performance of the JSRC. 

Since the proposed weighted scheme is solved using alternating minimization, a set of experiments are performed to test its performance sensitivity to the different initialization of the modalities weights $\mu^s$ and regularization parameters $\lambda_{\mu^s}$. In each experiment, all initial weights are set to a different value. Also all $\lambda_{\mu^s}$ are set to a different value, instead of being set by Eq.~(\ref{eq:WeightRegul}). The sparse coefficients $A$ are initialized to zero in all the experiments. We observed similar results for relatively large variations in initial weights and regularization parameters. The performance of the weighted scheme degrades if the regularization parameters are set to be too small. On the other hand, its performance approaches that of the non-weighted scheme for large values of the regularization parameters, as expected by observing cost function~(\ref{eq:WeightedScheme}). It is also observed that setting the regularization parameters by Eq.~(\ref{eq:WeightRegul}) with all the weights being initialized to $1/S$ persistently works well. 
                                  
\subsection{Multimodal target classification}\label{sec:Border}
This section presents the results of target classification on a field dataset consisting of measurements obtained from a passive infrared (PIR) and three seismic sensors of an unattended ground sensor system as discussed in~\cite{BRSDN13}. 
Symbolic dynamic filtering is used for feature extraction from time-series data~\cite{BRSDN13}. The subset of data used here consists of two days data. Day 1 includes 47 human targets and 35 animal-led-by-human targets while the corresponding numbers for Day 2 are 32 and 34, respectively. A two-way cross-validation is used to assess the performance of the classification algorithms, i.e. Day 1 data is used for training and Day 2 is used as test data and vice versa. 


For MTSRC, the tree-structured set of groups is selected to be $\mathcal{G} = \lbrace g_1, g_2, g_3, g_4, g_5, g_6 \rbrace= $ $\lbrace \lbrace 1 \rbrace, \lbrace 2 \rbrace, \lbrace 3 \rbrace, \lbrace 1,2,3 \rbrace, \lbrace 4 \rbrace, \lbrace 1,2,3,4 \rbrace \rbrace$ where $1$, $2$ and $3$ refer to the seismic channels and $4$ refers to the PIR channel. Table~\ref{tab:TargetClassResults} summarizes the average human detection rate (HDR), human false alarm rate (HFAR), and misclassification rates (MR) obtained using different multimodal classification algorithms. As seen, the proposed JSRC-W and MTSRC-W algorithms resulted in the best HDR and HFAR and, consequently the best overall performance. Moreover, if the different modalities are weighted equally, the MTSRC achieves the best performance. 

\begin{table}[!t] 
\caption{Results of target classification obtained by different multimodal classification algorithms by fusing 1 PIR and 3 seismic sensors data. HDR: Human Detection Rate, HFAR: Human False Alarm Rate, M: Misclassification.}
\label{tab:TargetClassResults}
\centering
\small
\begin{tabular}{cccc}
& HDR & HFAR & MR \\
\toprule
SVM-Sum & 0.94 & 0.15 & 10.61\% \\
HSRC  & 0.96 & 0.10 & 6.76\% \\
JDSRC & 0.97 & 0.09 & 8.11\% \\
RCR & 0.94 & 0.12 & 8.78\% \\
JSRC  & 1.00 & 0.12 & 5.41\% \\
JSRC-W  & 1.00 & 0.07 &  3.38\% \\
MTSRC & 1.00 & 0.09 & 4.05\% \\
MTSRC-W & 1.00 & 0.07 & 3.38\% \\
\bottomrule
\end{tabular}
\end{table} 
\begin{figure}[!t]
	\centering
         \includegraphics[scale=0.16]{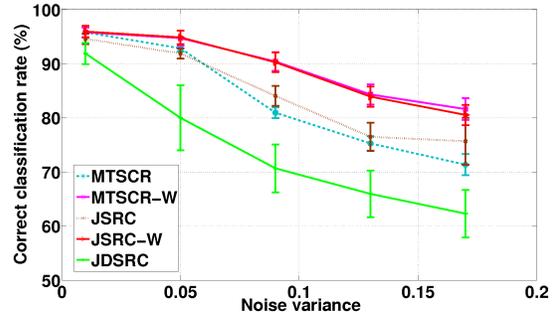}
         \caption{Correct classification rates of multimodal classification algorithms in dealing with random noise of varying variance that is added to the seismic 1 channel.}
      \label{fig:CCR_BorderNoise}
\end{figure}
\begin{figure}[!t]
		\centering
         \includegraphics[scale=0.16]{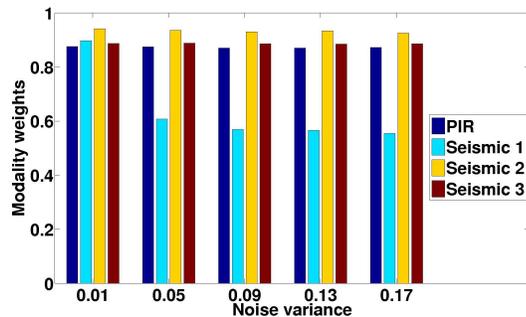}
         \caption{Averaged weights for each modality obtained by MTSRC-W algorithm when test samples from second modality, seismic sensor 1, are perturbed with random noise. As the noise level increases, the weight for second modality decreases while the corresponding weights for other unperturbed modalities remains almost constant.}
      \label{fig:BorderModalWeights}
\end{figure}

To evaluate the robustness of the proposed algorithms, random noise with varying variance is injected to the test samples of the seismic sensor 1 measurements. Fig.~\ref{fig:CCR_BorderNoise} shows the CCR obtained using different methods. The proposed weighted scheme has the most robust performance in dealing with noise. It is also seen that JSRC algorithm performs better than MTSRC as the level of noise increases. One possible reason is that in MTSRC the assumption of strong correlation between the three seismic channels are not valid with large value of noises. Fig.~\ref{fig:BorderModalWeights} shows averaged modality weights generated by the MTSRC-W. As expected, weight of the perturbed modality decreases as noise level increases. 

\section{Conclusions}\label{sec:Conclusions}

This paper presents the reformulation of tree-structured sparsity models for the purpose of multimodal classification and proposes an accelerated proximal gradient method to solve this class of problems. The tree-structured sparsity allows extraction of cross-correlated information among multiple modalities at different granularities. The paper also presents a possibilistic weighting scheme to jointly represent and quantify multimodal test samples by using several sparsity priors. This formulation provides a framework for robust fusion of available sources based on their respective reliability. The results show that the proposed algorithms achieve the state-of-the-art performances on the field data of three applications: multiview face recognition, multimodal face recognition, and multimodal target classification.

{
\renewcommand{\baselinestretch}{0.9}
\small
\bibliographystyle{ieee}
\bibliography{referencesAbbreviated}
\renewcommand{\baselinestretch}{1}
}

\end{document}